\RequirePackage{amsmath}
\documentclass[runningheads]{llncs}
\usepackage[english]{babel}
\usepackage[utf8]{inputenc}
\usepackage[T1]{fontenc}
\usepackage{amsmath, amssymb}
\usepackage{enumitem}
\usepackage{graphicx}
\usepackage{hyperref}
\usepackage{mathrsfs}
\usepackage{booktabs}
\usepackage{color}
\usepackage{algorithm}
\usepackage{algorithmic}

\DeclareMathOperator{\POS}{\text{POS}}
\DeclareMathOperator{\att}{\mathcal{A}}
\DeclareMathOperator{\tnorm}{\mathcal{T}}

\DeclareMathOperator{\implicator}{\mathcal{I}}

\begin{document}
\title{Optimising the attribute order in Fuzzy Rough Rule Induction}
%
%
\author{Henri Bollaert\inst{1}\orcidID{0000-0002-0664-2624} 
\and Chris Cornelis\inst{1}\orcidID{0000-0002-6852-4041}
\and Marko Palangetić\inst{1}\orcidID{0000-0002-6366-0634}
\and Salvatore Greco\inst{2}\orcidID{0000-0001-8293-8227}
\and Roman Słowiński\inst{3,4}\orcidID{0000-0002-5200-7795}
}
\authorrunning{H. Bollaert et al.}
%
\institute{
Department of Mathematics, Computer Science and Statistics, Ghent University, Ghent, Belgium
\email{henri.bollaert@ugent.be}\\
\and
Faculty of Economics, University of Catania, Italy
\and
Institute of Computing Science, Poznań University of Technology, 60-956 Poznań, Poland
\and
Systems Research Institute, Polish Academy of Sciences, 01-447 Warsaw, Poland
}
\maketitle              
\begin{abstract}

Interpretability is the next pivotal frontier in machine learning research. In the pursuit of glass box models --- as opposed to black box models, like random forests or neural networks --- rule induction algorithms are a logical and promising avenue, as the rules can easily be understood by humans. In \cite{BOLLAERT2025}, we introduced FRRI, a novel rule induction algorithm based on fuzzy rough set theory. We demonstrated experimentally that FRRI outperformed other rule induction methods with regards to accuracy and number of rules. FRRI leverages a fuzzy indiscernibility relation to partition the data space into fuzzy granules, which are then combined into a minimal covering set of rules. This indiscernibility relation is constructed by removing attributes from rules in a greedy way. This raises the question: does the order of the attributes matter? In this paper, we show that optimising only the order of attributes using known methods from fuzzy rough set theory and classical machine learning does not improve the performance of FRRI on multiple metrics. However, removing a small number of attributes using fuzzy rough feature selection during this step positively affects balanced accuracy and the average rule length.

\end{abstract}
\section{Introduction}
In a previous paper \cite{BOLLAERT2025}, we proposed a novel rule induction algorithm called Fuzzy-Rough Rule Induction (FRRI). FRRI combines the best aspects of fuzzy rule induction algorithms and rough rule induction algorithms to generate small sets of short rules that accurately summarise the training data and predict the classes of new objects. In the first step of the algorithm, which we will explain in more detail later, a greedy selection of attributes is used. When a greedy approach is used, it often pays dividends to reorder the input. For example, in bin packing algorithms, sorting the inputs in decreasing order results in considerable improvement in worst case behaviour \cite{JOHNSON1974272}. In this paper, we hypothesise that reordering the attributes in the rule shortening step would result in a smaller number of rules, which are also shorter and perform better at classification tasks.

The remainder of this paper is structured as follows. Section \ref{sec:prelim} recalls the required theoretical background, including a short introduction on fuzzy sets and rough sets. We give a brief explanation of the FRRI algorithm in Section \ref{sec:frri-expl} and of  the three considered feature selection methods in Section \ref{sec:feat_sel_overview}. Section \ref{sec:exp} describes the experiments that we set up to evaluate the effect of feature selection on FRRI in terms of prediction accuracy, number of generated rules and rule length. Finally, Section \ref{sec:conclusion} concludes the paper and outlines our future work.

\section{Preliminaries}\label{sec:prelim}

\subsection{Fuzzy sets}

In this paper, \(U\) will indicate a finite, non-empty set, called the \emph{universe}. A \emph{fuzzy set} \cite{fuzzysets} \(A\) in \(U\) is a function \(A \colon U \to [0,1]\). For \(u\) in \(U\), \(A(u)\) is called the membership degree of \(u\) in \(A\). The set of all fuzzy sets of $U$ is denoted as $\mathscr{F}(U)$. Let $A, B \in \mathscr{F}(U)$ be two fuzzy sets. $A$ is a fuzzy subset of $B$ if 
\[
(\forall u \in U)(A(u) \leq B(u))
\]
The union of a fuzzy set $A$ and a fuzzy set $B$ is the fuzzy set
\[
(\forall u \in U)((A \cup B)(u) = \max(A(u), B(u))
\]

In fuzzy logic and fuzzy set theory, we use \emph{triangular norms} and \emph{implicators} to generalise the classical connectives conjunction (\(\wedge\)) and implication (\(\rightarrow\)), respectively. A triangular norm or \emph{t-norm} is a function \(\tnorm \colon [0,1]^2 \to [0,1]\) which has \(1\) as neutral element, is commutative and associative and is increasing in both arguments. An \emph{implicator} is a function \(\implicator \colon [0,1]^2 \to [0,1]\) which is decreasing in the first argument and increasing in the second one. Additionally, the following boundary conditions should hold: 
\[\implicator(0,0) = \implicator(0,1) = \implicator(1,1) = 1 \text{ and } \implicator(1,0)=0.\]

A \emph{(binary) fuzzy relation} on a universe \(U\) is simply a fuzzy subset of \(U \times U\). A fuzzy relation \(R\) is \emph{reflexive} if \(R(u,u) = 1\) for all elements \(u \in U\). It is \emph{symmetric} if for every pair of elements \(u\), \(v\) in the universe, \(R(u, v) = R(v,u)\). For a given t-norm $\tnorm$, $R$ is \emph{$\tnorm$-transitive} if for every three elements \(u, v, w\), $\tnorm(R(u, v), R(v, w)) \leq R(u, w)$. A reflexive and $\tnorm$-transitive fuzzy relation is called a fuzzy \emph{$\tnorm$-preorder} relation. A fuzzy \emph{tolerance} relation is a reflexive and symmetric fuzzy relation. Given an element \(u \in U\), its \emph{fuzzy tolerance class} w.r.t.\ to a fuzzy tolerance relation \(R\) is the fuzzy set \(Ru (v) = R(u, v)\), for any \(v \in U\).

\subsection{Fuzzy rough sets}

In classical rough set theory \cite{roughsets}, we examine a universe divided into equivalence classes by an equivalence relation. We can then approximate a concept (i.e., a subset of \(U\)) using two crisp sets: the lower and the upper approximation. The \emph{lower approximation} contains all elements which are certainly part of the concept. It is defined as the union of the equivalence classes which are subsets of the concept. The \emph{upper approximation}, which contains all elements which might possibly be a part of the concept, is defined as the union of all equivalence classes which have a non-empty intersection with the concept in question. In fuzzy rough sets \cite{fuzzyroughsets}, we replace the crisp equivalence relation with a binary fuzzy relation \(R\) on \(U\) and, in turn, the upper and lower approximations also become fuzzy sets. Let \(A\) be a fuzzy set in \(U\). Furthermore, let \(\tnorm\) be a t-norm and \(\implicator\) be an implicator. We define the \emph{fuzzy rough set (FRS)} for the fuzzy set \(A\) w.r.t.\ the fuzzy relation \(R\) as the pair \((\underline{A}_R^{\implicator}, \overline{A}_R^{\tnorm})\), with, for \(u\) in \(U\):
\begin{itemize}
    \item The \emph{fuzzy lower approximation} \(\underline{A}_R^{\implicator}\) of $A$ is the fuzzy set
    \begin{equation}\label{eq:flowappr}        
        \underline{A}_R^{\implicator}(u) = \min_{v \in U} \implicator(R(u, v), A(v))
    \end{equation}
    \item The \emph{fuzzy upper approximation} \(\overline{A}_R^{\tnorm}\) is the fuzzy set
    \begin{equation}\label{eq:fuppappr}
        \overline{A}_R^{\tnorm}(u) = \max_{v \in U} \tnorm(R(u, v), A(v))
    \end{equation}
\end{itemize}

\subsection{Information and decision systems}\label{sec:infosys}

An \emph{information system} \cite{roughsets} is a pair \((U, \att)\), with \(U\)  a finite, non-empty \emph{universe} of objects, and \(\att\) a finite, non-empty set of \emph{attributes} describing these objects. Each attribute \(a \in \att\) has an associated domain of possible values \(V_a\) and an associated function \(a \colon U \to V_a\) which maps an object \(u \in U\) to its value \(a(u)\) for attribute \(a\). An attribute is \emph{numerical} if its value set is a closed real interval. 
In the remainder of this paper, we will only work in \emph{normalised} information systems with numerical condition attributes, where all attributes are scaled such that they have unit range, i.e., such that $V_a = [0,1]$ for all attributes $a \in \att$. This scaling step is important because the original range of a condition attribute often has no bearing on the true importance of the feature as a predictor for the class. To this aim, we will preprocess information systems by performing \emph{minimum-maximum normalisation}, i.e., we replace each original attribute function \(a \in \att\) with the following alternate attribute function $a'$:
\begin{align*}
	a' \colon &U \to [0, 1] \\
	&u \mapsto \frac{a(u) - \min \{a(v) \mid v \in U\}}{\max \{a(v) \mid v \in U\} - \min \{a(v) \mid v \in U\}}
\end{align*}
In Section \ref{sec:frri-expl}, we will sometimes look at novel objects that are not yet part of the information system under consideration. In such a case, it is possible that the value for a condition attribute $a$ of that object $v$ is outside of the range observed in the original information system. We solve this by setting $a'(v)$ to 1 if $a(v)$ is larger than $\max \{a(u) \mid u \in U\}$, and setting it to 0 if $a(v)$ is smaller than $\min \{a(u) \mid u \in U\}$.

In the following sections, we will assume that each condition attribute $a$ has been normalised, and stop writing $a'$.

If there is no inherent order on the values of an attribute --- as might often by the case with attributes like age or height --- they can be compared by means of a binary fuzzy \emph{indiscernibility} relation \(R_i\) defined on \(U \times U\). In this paper, we will use the following relation:
\begin{equation}\label{eq:indiscrel}
    R_i(u, v) = 1 - |a(u) - a(v)|
\end{equation}
However, imagine a case where an attribute represents length, which does have an inherent order. In such cases, we can use a fuzzy \emph{dominance} relation $R_d$, which is a fuzzy $\tnorm$-preorder relation for a given t-norm $\tnorm$ that encodes how much the first argument dominates the second. In this paper, we will use the following relation:
\begin{equation}
    R_d(u, v) = \min\left(1 - (a(v) - a(u)), 1\right)
\end{equation}
We can compare two objects in $U$ w.r.t.\ a subset \(B\) of \(\att\) using the fuzzy \(B\)-indiscernibility relation \(R_B\), defined as 
\begin{equation}
    R_B(u, v) = \tnorm(\underbrace{R_a(u, v)}_{a \in B})
\end{equation}
for each \(u\) and \(v\) in \(U\) and for a given t-norm \(\tnorm\), where $R_a$ is the relation used for comparing the values of attribute $a$. In general, \(R_B\) is a reflexive fuzzy relation.

A \emph{decision system} \((U, \att \cup \{d\})\) with a single \emph{decision attribute} \(d\) is an information system which makes a distinction between the \emph{condition} attributes \(\att\) and the decision attribute \(d\), which is not an element of \(\att\). In this paper, the decision attribute is assumed to be \emph{categorical}, which means that it has a finite, unordered value domain. We will call the values in \(V_d\), \emph{decision classes}. 
For a given object $u$ in $U$, we will identify its decision class $d(u)$ with the set of all objects that have the same value for the decision attribute.

Given a subset of attributes \(B \subseteq \att\), the fuzzy \(B\)-positive region is a fuzzy set in the universe \(U\) that contains each object \(u\) to the extent that all objects which are approximately similar to \(u\) w.r.t.\ the attributes in \(B\), have the same decision value \(d\) \cite{CORNELIS2010209}:
\[
\POS_B(u) = \left(\bigcup_{c \in V_d} \underline{c}_{R_B}^{\implicator} \right) (u)
\]
It is calculated as the membership degree of $u$ to the union of the fuzzy lower approximations of all decision classes of the decision system, which in this case reduces to
\[
\POS_B(u) = \underline{d(u)}_{R_B}^{\implicator}(u)
\]
as an object $u$ cannot belong to the lower approximation of any other class.

The predictive ability of the attributes in \( B \) with respect to the decision attribute $d$ is reflected by the \emph{degree of dependency} \( \gamma_B \) of \( d \) on \( B \), defined as the ratio
\begin{equation}
    \gamma_B = \frac{|POS_B|}{|POS_A|} = \frac{\sum\limits_{u \in U} POS_B(u)}{\sum\limits_{u \in U} POS_A(u)}
\end{equation}
where the cardinality of a fuzzy set is defined by means of the well-known sigma count.

\section{An overview of the Fuzzy Rough Rule Induction algorithm}\label{sec:frri-expl}

Fuzzy-Rough Rule Induction (FRRI) \cite{BOLLAERT2025} is a novel rule induction algorithm that integrates concepts from fuzzy set theory and rough set theory to construct interpretable decision rules. The algorithm operates in two primary phases: rule shortening and rule selection. Before we summarise these steps, it is important to detail the format of the rules used in FRRI.

\subsection{Rule format}

A rule in FRRI takes the following form:
\begin{equation*}
        \textsc{if} \text{ antecedent } \textsc{then} \text{ consequent}
\end{equation*}
where the antecedent consists of a conjunction of conditions related to the attributes of the original training instance, and the consequent corresponds to its class label. The antecedent conditions can be categorised as follows:
\begin{itemize}
    \item \textsc{similar}: The attribute value is similar to the observed value in the training object. This is modeled using the fuzzy indiscernibility relation $R_i$.
    \item \textsc{dominant}: The attribute value is less than or similar to the observed value. This is modeled using the fuzzy dominance relation $R_d$.
    \item \textsc{dominated}: The attribute value is greater than or similar to the observed value.  This is also modeled using the fuzzy dominance relation $R_d$, but with the position of the arguments swapped.
\end{itemize}
Additionally, it is also possible that an attribute is not used in a rule. The antecedent of a rule $r$ derived from an object $u_r$ results in the \emph{matching degree} of another object $v$ to $r$ as
\begin{equation*}
    M_r(v) = \min\left\langle R_{\textsc{type}_r(a)}(a(u_r), a(v)) \mid a \in \att \right\rangle
\end{equation*}
The antecedent and consequent of $r$ are then aggregated with the minimum t-norm to calculate the covering set with membership function:
 \begin{equation*}
    S_r(v) = \min\left(M_r(v), \underline{d(u_r)}^{\implicator}_{R_{\att}}(u_r)\right)
\end{equation*}

\subsection{Rule shortening}
FRRI starts from the complete set $\mathscr{R}_t$ of total rules, each derived from a single object in the training set. A total rule for an object $u$ is a rule where the antecedent contains a \textsc{similar} condition for every attribute.

To minimise redundancy and improve generalisation, the rule shortening phase removes unnecessary conditions from a rule while maintaining its discriminative power. This process ensures that each rule remains as general as possible without covering instances from other classes. This process is applied to each total rule independently in a greedy fashion as follows. 

We consider each of the attributes of the total rule, one after the other. For each attribute $a_i$ for an object $u$, we consecutively examine the possibility of setting the type of the condition corresponding to $a_i$ for the rule corresponding to $u$ as \textsc{unused, dominant, dominated} or \textsc{similar}. If for a type $t$, $S_{r^*(u)}$ is a subset of $d(u)$, then we set the type of $a_i$ to that type $t$ for this object. If not, we continue to the next type in the list.
We repeat this process for each attribute in the order $a_1, a_2, \dots, a_n$, and we repeat this \textsc{rule\_prune} procedure for each object in the training set. The result is a shortened, initial ruleset
\[\widehat{\mathscr{R}} = \{\hat{r}(u) \mid u \in U\}\]

\subsection{Rule selection}\label{sec:rule-selection}
Once the shortened ruleset is generated, FRRI selects a minimal subset that still covers the entire training dataset. This is formulated as an integer programming problem, where the objective is to minimise the number of selected rules while ensuring that all training instances remain covered. This optimisation problem can be expressed as:

\begin{align*}
    \text{Minimise: } &\sum_{u \in U} r_u \\
    \text{subject to } &(\forall v \in U)\left(\sum_{u \in U}r_u \times z_{u, v} \geq 1 \right)
\end{align*}
where $z_{u,v} = 1$ if rule $\hat{r}(u)$ covers instance $v$ and 0 otherwise. $r_u$ is a binary variable indicating whether rule $\hat{r}(u)$ is included in the final ruleset $\mathscr{R}$.

\subsection{Inference}
The resulting ruleset can be used to classify new objects. The class of a new instance is determined by computing the covering degree of the instance for each rule in the final ruleset. The instance is then assigned to the class that corresponds to the consequent of the rule with the highest covering degree.

\section{Feature selection}\label{sec:feat_sel_overview}

It is often the case that not all of the condition attributes in a decision system are relevant to the decision attribute. In such occasions, feature selection or feature extraction procedures are often used \cite{feat-sel-review2024}. Techniques from the latter type produce new features from the original data using a transformation. Procedures from the former type select a subset of attributes, $\att' \subset \att$, which are then used instead of the complete set $\att$. Three types of feature selection algorithms exists: filter, wrapper and embedded methods \cite{feat-sel-review2024}. For our application, we are only interested in filter methods, which are used as a preprocessing step, separate from any other algorithms that might use the decision system as input. We will consider three different filter methods: QuickReduct, mutual information-based feature selection and correlation-based feature selection, which we will cover in more detail shortly. Each of these methods will be used to create a new ordered list of attributes, from most important to least important according to their respective criteria, as a preprocessing step before the rule reduction step explained in Section \ref{sec:rule-selection}.

\subsection{Fuzzy rough feature selection}\label{sec:frfs}

Fuzzy rough feature selection (FRFS) is usually implemented by means of the QuickReduct algorithm \cite{quickreduct}. This is a greedy algorithm with a hillclimbing heuristic designed to find a set of attributes \(B\) for which \(\gamma_B\) is 1. In this case, \(B\) is called a decision superreduct. It is a bottom-up algorithm which keeps track of a single set of attributes \(B'\), and iteratively adds the attribute \(a \in \att\) for which \(\gamma_{B' \cup \{a\}}\) is maximal, until a superreduct is obtained. The pseudocode can be found in Algorithm \ref{alg:quickreduct}. The output of this algorithm is also dependent on the order of attributes during the iteration process.

\begin{algorithm}[hbt]
    \caption{The \textsc{QuickReduct} procedure}
    \label{alg:quickreduct}
    \begin{algorithmic}[1]
       \STATE \( B \leftarrow \{\} \)
        \REPEAT
        \STATE \( T \leftarrow B \)
        \FORALL{ \( a \in (\mathcal{A} \setminus B) \) }
            \IF{ \( \gamma_{B \cup \{a\}} > \gamma_T \) }
                \STATE \( T \leftarrow B \cup \{a\} \)
            \ENDIF
        \ENDFOR
        \STATE \( B \leftarrow T \)
    \UNTIL{ \( \gamma_B = \gamma_A \) }
    \RETURN \( B \)
    \end{algorithmic}
\end{algorithm}

 As QuickReduct was originally designed to only construct superreducts, the algorithm does not always order the entire set of attributes. If this happens, we add the remaining attributes in their original order, as given by the universe under consideration.

 \subsection{Scoring-based feature selection}

Scoring-based feature selection methods give each attribute a score and select the attributes with the highest scores. In this paper, the scores we consider are the Pearson correlation score and the mutual information score between the condition attribute and the decision attribute. We calculate these scores for every condition attribute and then sort them in descending order.


\section{Experimental evaluation}\label{sec:exp}

\subsection{Methods and evaluation measures}

\begin{table}[ht]
\centering
\caption{Details of the benchmark datasets used for these experiments: abbreviated name, the number of features of each dataset ($|\att|$), the number of instances ($|U|$) and the number of classes ($|V_d|$).}
\begin{tabular}{llrrrr}
\toprule
dataset         & abbr.        & \(|\att|\) & \(|U|\) & \(|V_d|\)   \\
\midrule
australian      & aus          & 14      & 690     & 2  \\
bands           & bands        & 19      & 365     & 2  \\
bupa            & bupa         & 6       & 345     & 2  \\
cleveland       & cleve        & 13      & 297     & 5  \\
dermatology     & derma        & 34      & 358     & 6  \\
ecoli           & ecoli        & 7       & 336     & 8  \\
glass           & glass        & 9       & 214     & 6  \\
heart           & heart        & 13      & 270     & 2  \\
ionosphere      & iono         & 33      & 351     & 2  \\
pima            & pima         & 8       & 768     & 2  \\
sonar           & sonar        & 60      & 208     & 2  \\
spectfheart     & spect        & 44      & 267     & 2  \\
vehicle         & vehi         & 18      & 846     & 4  \\
vowel           & vowel        & 13      & 990     & 11 \\
wine            & wine         & 13      & 178     & 3  \\
winequality-red & red          & 11      & 1599    & 6  \\
wisconsin       & wisc         & 9       & 683     & 2  \\
yeast           & yeast        & 8       & 1484    & 10 \\ 
\bottomrule
\end{tabular}
\label{tab:datasets}
\end{table}

We perform a computational experiment to evaluate the impact of a feature selection preprocessing step on the performance of the FRRI algorithm on 18 benchmark numerical datasets from the KEEL dataset repository\footnote{Some of these datasets have categorical condition features. We opted to remove these features from the dataset before using them in our experiments.} \cite{keeldatarepo}. The details of these datasets can be found in Table \ref{tab:datasets}, where we list the number of numerical features of each dataset ($|A|$), as well as the number of instances ($|U|$) and classes ($|V_d|$). 
We also list an abbreviated name for each dataset.

We use our own Python implementation of FRRI, which uses the \textsc{gurobipy} package for Python \cite{gurobi} to solve the optimisation problem in the rule selection step. For FRFS, we build on the implementation in the \texttt{fuzzy-rough-learn} Python package \cite{lenz2020fuzzy,lenz22fuzzyroughlearn}. We add the possibility to obtain the order in which the attributes are selected in the FRFS procedure. We denote this with \texttt{ofrfs}, where the O stands for ordered.
Additionally, for mutual information and correlation, we use the implementations from the scikit-learn package \cite{scikit-learn}. 
We denote these as \texttt{mi} and \texttt{pcc} respectively.
We compare the following versions of FRRI, where \texttt{***} is a placeholder for \texttt{ofrfs}, \texttt{mi}, or \texttt{pcc}:
\begin{itemize}
    \item \texttt{control}, which is the original version of FRRI with the default parameters.
    \item \texttt{***-1}, where we only change the order of the attributes with a feature selection method but do not reduce their number.
    \item \texttt{***-.x}, where we change the order and retain $\left\lfloor \frac{x}{10} \times |\att| \right\rfloor$ attributes. However, if this amount is 1 or 0, we default to the next method to avoid trivial cases without any attributes.
    \item \texttt{ofrfs-0}, where we stop adding attributes when the degree of dependency of the selected set is equal to 1.
\end{itemize}
We use the default parameters for all algorithms, unless otherwise specified above.

For each algorithm, we perform ten-fold cross-validation on each dataset, using the same folds, which are provided in the KEEL repository, for each algorithm, after which we calculate the average balanced accuracy of the algorithm on the ten training folds. The balanced accuracy is the arithmetic mean of the recall on each class. Moreover, we also calculate the average number of rules and the average rule length over each ruleset and over each of the ten folds of each dataset. 
To determine the statistical significance of our results, we apply a two-stage procedure, which starts with a Friedman test \cite{multiple-comp-friedman}. If the result of that test is significant, we continue with a Conover post-hoc pairwise comparison procedure \cite{multiple-comp-conover,conover_test_big_citations} to determine the location of the significant differences. 
We use the Wilcoxon test \cite{pairwise-comp-wilcoxon} to detect differences between two methods.

\subsection{Results}

\begin{figure}[ht]
    \centering
    \caption{The average balanced accuracy of \texttt{ofrfs} with an increasing number of retained attributes over all datasets.}
    \includegraphics[width=0.8\linewidth]{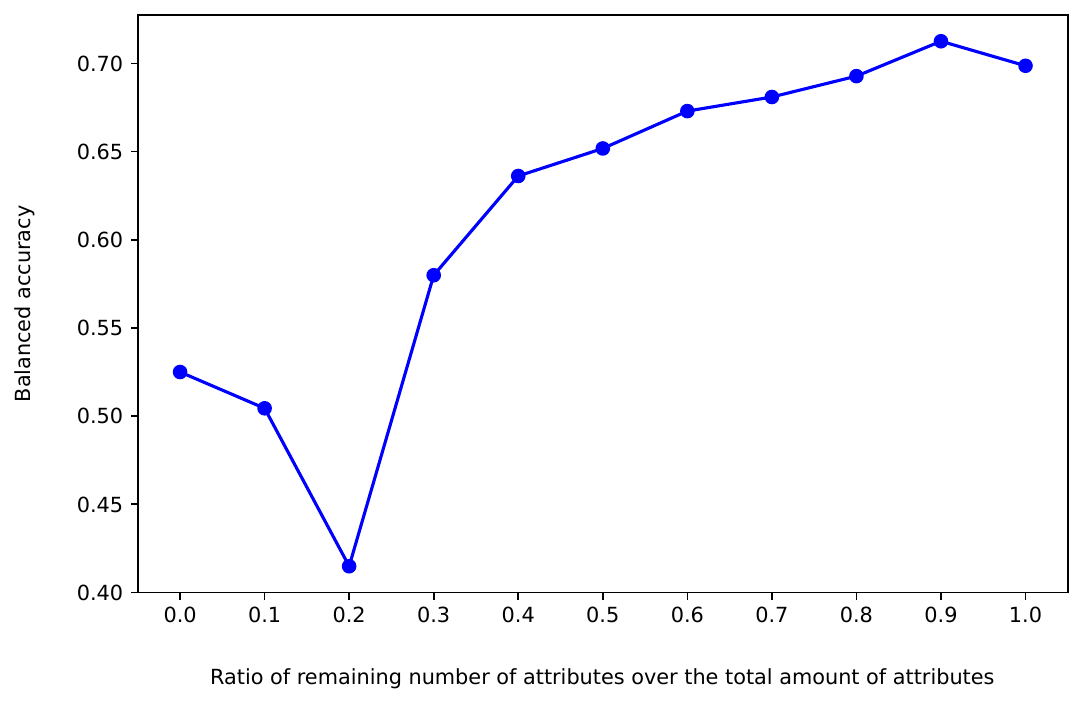}
    \label{fig:bal-acc}
\end{figure}

\begin{table}[hp]
    \centering
    \caption{Balanced accuracy of FRRI with the three feature selection methods as preprocessing on the benchmark datasets.}
    \begin{tabular}{lrrrrrrrrrrr}
\toprule
{} &         {control} &         \texttt{ofrfs-1} &       \text{ofrfs-0.9} &       \text{ofrfs-0.8} & \texttt{ofrfs-0} &            \texttt{mi-1} &          \texttt{mi-0.9} & \texttt{mi-0.8} &  \texttt{pcc-1} & \texttt{pcc-0.9} &         \texttt{pcc-0.8} \\
\midrule
aus      &  \textbf{0.825} &           0.813 &           0.816 &           0.786 &   0.583 &           0.805 &           0.800 &  0.798 &  0.805 &   0.811 &           0.798 \\
bands           &           0.665 &  \textbf{0.693} &           0.691 &           0.632 &   0.568 &           0.642 &           0.654 &  0.641 &  0.669 &   0.655 &           0.632 \\
bupa            &           0.593 &           0.578 &           0.611 &           0.611 &   0.537 &           0.577 &  \textbf{0.621} &  0.603 &  0.603 &   0.598 &           0.598 \\
cleve       &  \textbf{0.318} &           0.308 &           0.305 &           0.267 &   0.221 &           0.285 &           0.287 &  0.297 &  0.287 &   0.273 &           0.266 \\
derma     &           0.903 &           0.894 &           0.920 &           0.909 &   0.184 &           0.895 &           0.920 &  0.899 &  0.898 &   0.910 &  \textbf{0.920} \\
ecoli           &  \textbf{0.701} &           0.700 &           0.662 &           0.662 &   0.503 &           0.694 &           0.659 &  0.675 &  0.697 &   0.628 &           0.628 \\
glass           &           0.668 &           0.655 &  \textbf{0.718} &           0.675 &   0.589 &           0.673 &           0.672 &  0.681 &  0.646 &   0.676 &           0.563 \\
heart           &           0.731 &           0.748 &  \textbf{0.773} &           0.693 &   0.537 &           0.745 &           0.773 &  0.767 &  0.738 &   0.734 &           0.750 \\
iono      &           0.911 &           0.902 &           0.922 &  \textbf{0.938} &   0.839 &           0.922 &           0.919 &  0.910 &  0.906 &   0.908 &           0.925 \\
pima            &  \textbf{0.681} &           0.672 &           0.660 &           0.654 &   0.569 &           0.653 &           0.630 &  0.645 &  0.647 &   0.657 &           0.645 \\
sonar           &  \textbf{0.776} &           0.761 &           0.767 &           0.775 &   0.510 &           0.720 &           0.771 &  0.747 &  0.734 &   0.737 &           0.774 \\
spect     &           0.638 &           0.618 &  \textbf{0.668} &           0.645 &   0.509 &           0.659 &           0.640 &  0.645 &  0.650 &   0.610 &           0.572 \\
vehi         &           0.660 &           0.649 &           0.661 &  \textbf{0.694} &   0.481 &           0.641 &           0.693 &  0.679 &  0.648 &   0.652 &           0.656 \\
vowel           &           0.882 &           0.891 &           0.905 &           0.886 &   0.477 &  \textbf{0.914} &           0.902 &  0.894 &  0.910 &   0.888 &           0.859 \\
wine            &  \textbf{0.949} &           0.949 &           0.944 &           0.902 &   0.789 &           0.918 &           0.931 &  0.919 &  0.943 &   0.919 &           0.935 \\
red &           0.323 &           0.331 &  \textbf{0.376} &           0.339 &   0.301 &           0.330 &           0.359 &  0.334 &  0.337 &   0.323 &           0.326 \\
wisc       &           0.943 &           0.947 &           0.947 &           0.943 &   0.937 &           0.943 &           0.931 &  0.930 &  0.947 &   0.928 &  \textbf{0.949} \\
yeast           &  \textbf{0.483} &           0.469 &           0.480 &           0.460 &   0.314 &           0.467 &           0.411 &  0.414 &  0.473 &   0.355 &           0.288 \\
\midrule
mean            &           0.703 &           0.699 &  \textbf{0.713} &           0.693 &   0.525 &           0.694 &           0.699 &  0.693 &  0.696 &   0.681 &           0.671 \\
\bottomrule
\end{tabular}
    \label{tab:bal_acc}
\end{table}

Table \ref{tab:bal_acc} contains the balanced accuracy of the algorithms under consideration on the benchmark datasets. We only show the results of when keeping at least $90\%$ and $80\%$ of the attributes, as removing even more results in an unacceptable decrease in the performance of FRRI, as shown in Figure \ref{fig:bal-acc}.
We see that on average, \texttt{ofrfs} with $90\%$ of the attributes performs the best. It is also tied with \texttt{control} for the most (7) datasets where it achieves the highest accuracy out of the \texttt{ofrfs} algorithms. A Friedman test shows a significant difference between the control and \texttt{ofrfs} algorithms ($p < 10^{-7}$). The post-hoc tests show that only \texttt{ofrfs-0} is significantly worse than all other algorithms, and that \texttt{ofrfs-0.9} is significantly better than \texttt{pcc-0.9} and \texttt{pcc-0.8}. With a $p$-value of $0.071$, a one-sided Wilcoxon test to determine whether \texttt{ofrfs-.9} is more accurate than the \texttt{control} is weakly significant. For all three feature selection methods, we see a decrease in accuracy as we retain less attributes, however, for both \texttt{ofrfs} and \texttt{mi}, we see the best accuracy with up to $10\%$ of the attributes removed.

\begin{table}[hp]
    \centering
    \caption{The average number of rules of FRRI with the three feature selection methods as preprocessing on the benchmark datasets.}
    \begin{tabular}{lrrrrrrrrrrr}
\toprule
{} &         {control} &         \texttt{ofrfs-1} &       \text{ofrfs-0.9} &       \text{ofrfs-0.8} & \texttt{ofrfs-0} &            \texttt{mi-1} &          \texttt{mi-0.9} & \texttt{mi-0.8} &  \texttt{pcc-1} & \texttt{pcc-0.9} &         \texttt{pcc-0.8} \\
\midrule
aust      &   \textbf{93} &            94 &           95 &           100 &     294 &           93 &           94 &     98 &            94 &      97 &      97 \\
bands           &            57 &   \textbf{56} &           57 &            59 &     228 &           58 &           61 &     59 &            60 &      64 &      66 \\
bupa            &            77 &            77 &           79 &            79 &     165 &  \textbf{76} &           88 &     86 &            79 &      87 &      87 \\
cleve       &           104 &           104 &          105 &           109 &     149 &          100 &          104 &    105 &   \textbf{99} &     102 &     106 \\
derma     &            21 &            21 &  \textbf{18} &            21 &     320 &           23 &           24 &     22 &            20 &      22 &      23 \\
ecoli           &   \textbf{55} &            55 &           63 &            63 &      88 &           56 &           57 &     56 &            55 &      59 &      59 \\
glass           &            52 &            51 &  \textbf{45} &            46 &      59 &           51 &           47 &     46 &            50 &      50 &      59 \\
heart           &            49 &            49 &           48 &            52 &     140 &           44 &           44 &     43 &   \textbf{43} &      44 &      46 \\
iono      &            20 &            19 &           19 &   \textbf{19} &      62 &           20 &           22 &     22 &   \textbf{19} &      19 &      20 \\
pima            &  \textbf{119} &           121 &          130 &           150 &     261 &          124 &          134 &    141 &           125 &     133 &     140 \\
sonar           &   \textbf{17} &            17 &           19 &            19 &      82 &           19 &           20 &     20 &            18 &      20 &      21 \\
spect     &            24 &            24 &           24 &            26 &      74 &           25 &           25 &     26 &   \textbf{24} &      24 &      28 \\
vehi         &           179 &           177 &          165 &  \textbf{150} &     388 &          183 &          152 &    167 &           154 &     156 &     159 \\
vowel           &           124 &           126 &          127 &           130 &     504 &          133 &          131 &    122 &  \textbf{120} &     138 &     156 \\
wine            &             7 &             8 &            9 &            11 &      40 &            9 &            8 &      8 &    \textbf{7} &       8 &       9 \\
red &           400 &           398 &          406 &           434 &     691 &          396 &          400 &    406 &  \textbf{392} &     406 &     424 \\
wisc       &            23 &            22 &           23 &            24 &      60 &           23 &  \textbf{22} &     25 &            23 &      23 &      23 \\
yeast           &           480 &  \textbf{480} &          532 &           580 &     764 &          486 &          497 &    517 &           485 &     647 &     748 \\
\midrule
mean            &           106 &           106 &          109 &           115 &     243 &          107 &          107 &    109 &  \textbf{104} &     116 &     126 \\
\bottomrule
\end{tabular}
    \label{tab:num_rules}
\end{table}

Let us consider now the impact of the feature ordering and reduction on the number of rules generated by FRRI. The average number for each dataset and each algorithm can be found in Table \ref{tab:num_rules}. We see that reducing the number of attributes in the input of FRRI results in a significant (Friedman $p$-value $< 10^{-9}$) increase of the number of generated rules, especially as we greatly reduce the number of attributes available to the rule induction algorithm. When we perform post-hoc tests, the significance only appears with a reduction of at least $20\%$ and increases thereafter. These post-hoc tests show no significant pairwise difference between any of the three feature selection methods when the same number of features are retained.

This trend is reversed when we look at the number of attributes per rule, which are shown in Table \ref{tab:rule_lengths}. This was expected. A one-sided Wilcoxon test shows that by reducing the number of attributes by $10\%$ with \texttt{ofrfs}, we get a significant decrease in the final average length of a rule in the ruleset generated by FRRI. This significance is not seen with \texttt{pcc} or \texttt{mi} when the same number of attributes is retained. However, further decreases in number of attributes given in the input of FRRI result in a significant decrease in the rule lengths, but, as seen before, this decrease is paired with a much too large decrease in predictive accuracy and a large growth of the number of rules.

In summary, the findings indicate that merely reordering the attributes does not significantly affect the performance of FRRI. The results suggest that the greedy algorithm used in the rule reduction of FRRI effectively identifies the most relevant attributes to represent an object, regardless of their initial order. In contrast, removing certain attributes does have a noticeable impact. Specifically, eliminating only a few attributes with FRFS specifically --- presumably those that are superfluous or redundant --- appears to be the most effective strategy. This selective and small-scale removal enhances FRRI's ability to identify the critical attributes for each object separately, whereas removing too many attributes with feature selection before FRRI might eliminate input features that would have been valuable selections for particular objects. The other two feature selection algorithms under consideration show similar trends to FRFS with regards to accuracy, rule set size and rule length, but with an overall lower performance, with no improvements gained over the default version of FRRI.

We offer the following possible explanation as to why a removing more attributes before FRRI is inefficient. Unlike FRRI, which removes attributes solely for the object under consideration, the examined feature selection methods eliminate attributes for every object at the same time. This distinction is important because, while a dataset may contain redundant attributes overall, different objects may require distinct subsets of attributes to be accurately distinguished. As such, the feature selection methods might remove attributes that are irrelevant to the dataset as a whole but which are crucial for discerning a particular object from its neighbours.

\begin{table}[htb]
    \centering
    \caption{The average rule lengths in the rulesets of FRRI with the three feature selection methods as preprocessing on the benchmark datasets.}
   \begin{tabular}{lrrrrrrrrrrr}
\toprule
{} &         {control} &         \texttt{ofrfs-1} &       \text{ofrfs-0.9} &       \text{ofrfs-0.8} & \texttt{ofrfs-0} &            \texttt{mi-1} &          \texttt{mi-0.9} & \texttt{mi-0.8} &  \texttt{pcc-1} & \texttt{pcc-0.9} &         \texttt{pcc-0.8} \\
\midrule
aust      &    5.16 &    5.34 &      5.33 &      5.24 &  \textbf{1.59} &  5.28 &   5.00 &   4.53 &  5.34 &    5.13 &    4.49 \\
bands           &    6.45 &    6.54 &      6.47 &      6.27 &  \textbf{0.57} &  6.45 &   6.29 &   6.02 &  6.19 &    6.18 &    5.57 \\
bupa            &    4.24 &    4.24 &      3.89 &      3.89 &  \textbf{1.58} &  4.21 &   3.77 &   3.74 &  4.07 &    3.64 &    3.64 \\
cleve       &    5.27 &    5.28 &      5.31 &      5.55 &  \textbf{2.52} &  5.62 &   5.46 &   5.12 &  5.67 &    5.63 &    5.27 \\
derma     &    4.38 &    4.38 &      4.33 &      4.96 &  \textbf{0.02} &  5.18 &   4.51 &   4.44 &  4.57 &    4.65 &    4.95 \\
ecoli           &    3.87 &    3.85 &      3.49 &      3.49 &  \textbf{2.77} &  3.86 &   3.85 &   3.86 &  3.91 &    3.76 &    3.76 \\
glass           &    4.14 &    4.10 &      3.93 &      3.81 &  \textbf{3.12} &  4.11 &   3.96 &   3.78 &  3.87 &    3.98 &    3.78 \\
heart           &    4.73 &    4.72 &      4.93 &      5.02 &  \textbf{1.45} &  5.15 &   4.91 &   4.52 &  5.05 &    5.17 &    4.74 \\
iono      &    4.49 &    4.55 &      4.34 &      4.20 &  \textbf{1.86} &  4.69 &   4.97 &   4.77 &  4.66 &    4.40 &    4.35 \\
pima            &    5.15 &    5.13 &      4.75 &      4.30 &  \textbf{2.56} &  5.06 &   4.82 &   4.40 &  5.12 &    4.82 &    4.44 \\
sonar           &    8.07 &    8.17 &      7.62 &      6.74 &  \textbf{1.95} &  8.05 &   7.83 &   7.30 &  7.60 &    7.78 &    7.68 \\
spect     &    8.05 &    8.05 &      8.30 &      8.12 &  \textbf{2.51} &  8.46 &   7.88 &   7.85 &  8.28 &    8.31 &    7.73 \\
vehi         &    5.40 &    5.34 &      5.20 &      5.06 &  \textbf{2.70} &  5.36 &   5.03 &   4.91 &  5.75 &    5.73 &    5.67 \\
vowel           &    6.57 &    6.51 &      6.24 &      5.59 &  \textbf{1.94} &  6.27 &   6.37 &   6.29 &  6.06 &    6.05 &    5.54 \\
wine            &    4.41 &    4.44 &      4.21 &      4.17 &  \textbf{1.92} &  4.55 &   4.63 &   4.17 &  4.21 &    4.17 &    4.45 \\
red &    5.83 &    5.92 &      5.79 &      5.43 &  \textbf{2.74} &  5.97 &   5.73 &   5.51 &  6.00 &    5.70 &    5.45 \\
wisc       &    3.85 &    3.68 &      3.56 &      3.45 &  \textbf{1.43} &  3.77 &   3.59 &   3.43 &  3.86 &    3.70 &    3.48 \\
yeast           &    4.94 &    4.93 &      4.44 &      3.72 &  \textbf{2.74} &  4.89 &   4.75 &   4.49 &  4.77 &    4.12 &    3.34 \\
\midrule
mean            &    5.28 &    5.29 &      5.12 &      4.95 &  \textbf{2.00} &  5.38 &   5.19 &   4.95 &  5.28 &    5.16 &    4.91 \\
\bottomrule
\end{tabular}

    \label{tab:rule_lengths}
\end{table}

\section{Conclusion and future work}\label{sec:conclusion}

In this article, we examined the impact of reordering and reducing the attributes using three feature selection methods: fuzzy rough feature selection, mutual information-based feature selection and correlation-based feature selection on the performance of the FRRI algorithm with respect to three key metrics: balanced accuracy, average ruleset size and average rule length. Experimental evaluation showed that a simple reordering with any of the three methods does not have a significant impact on any of these metrics. However, if we remove up to $10\%$ of the attributes with FRFS as well as reorder them, we see an improved balanced accuracy with significantly shorter rules in the majority of benchmark datasets. The impact on ruleset size is minimal. This gain is not seen with the other two methods. Additionally, more drastic attribute reduction has a negative effect on balanced accuracy. To conclude, we recommend users to preprocess their data with FRFS to remove up to $10\%$ of attributes and reorder the remainder before applying the FRRI algorithm, as this can result in significantly shorter rules that are more accurate, without a significant increase in the size of the ruleset.

We have identified some challenges for future work:
\begin{enumerate}
    \item We want to adapt our algorithm to regression and ordinal classification problems.
    \item We will reformulate the rule shortening step of FRRI as an optimisation problem, that we solve exactly, circumventing the need to use an attribute ordering pre-processing step. 
    \item We want to apply evolutionary fuzzy algorithms \cite{FERNANDEZ2015109} to the ruleset produced by FRRI to further increase test accuracy and reduce overfitting, which was recently applied successfully to Michigan-style fuzzy rules \cite{hinojosa2021improving}.
    \item We will evaluate the performance of FRRI on more challenging datasets.
    \item We also want to construct hierarchical rulesets, by combining similar rules into a higher-level, more general rule, thus reducing the number of rules and improving the explainablility.
\end{enumerate}

\subsubsection{\ackname} Henri Bollaert would like to thank the Special Research Fund of Ghent University (BOF-UGent) for funding his research (Grant No. BOF22/DOC/113) and the Fonds Professor Frans Wuytack in particular for supporting his participation in this conference. The research of Roman Słowiński was funded by the National Science Centre, Poland, under the MAESTRO programme (Grant No. 2023/50/A/HS4/00499).
%
%

\bibliographystyle{splncs04}
\bibliography{references}

\end{document}